\documentclass{styles/svproc}
\usepackage{url}

\usepackage{subfig}
\usepackage{graphicx}
\usepackage{amsmath}
\usepackage{cleveref}
\usepackage{booktabs} 
\usepackage{tikz}
\usepackage{tikz-3dplot}
\usepackage{pgfplots}
\usepackage{float}
\usepackage[product-units=single,per-mode=symbol,range-units=single,detect-all]{siunitx}
\pgfplotsset{compat=1.14}
\usetikzlibrary{
    calc,
    patterns,
    circuits.logic.US,
    circuits.ee.IEC,
    intersections,
    angles,
    quotes,
    arrows,
    shapes,
    graphs,
    positioning,
    }

\begin{document}
\mainmatter              
\title{Perception-aware Exploration for Consumer-grade UAVs}
\titlerunning{Perception-aware Exploration for Consumer-grade UAVs} 
\author{Svetlana Seliunina \and Daniel Schleich \and Sven Behnke}
\authorrunning{Svetlana Seliunina et al.} 
\institute{Autonomous Intelligent Systems group, Computer Science Institute VI – Intelligent Systems and Robotics, University of Bonn \\                          \email{seliunina@ais.uni-bonn.de}}

\maketitle              

\begin{abstract}
    In our work, we extend the current state-of-the-art approach for autonomous multi-UAV exploration to consumer-level UAVs, such as the DJI Mini 3 Pro. We propose a pipeline that selects viewpoint pairs from which the depth can be estimated and plans the trajectory that satisfies motion constraints necessary for odometry estimation. For the multi-UAV exploration, we propose a semi-distributed communication scheme that distributes the workload in a balanced manner. We evaluate our model performance in simulation for different numbers of UAVs and prove its ability to safely explore the environment and reconstruct the map even with the hardware limitations of consumer-grade UAVs. 
\end{abstract}
\section{Introduction}
Autonomous exploration is a rapidly developing field with numerous applications, and multi-UAV exploration holds significant potential. The use of several coordinating robots allows for a faster exploration of big environments, especially considering the limited battery life of a single quadrotor. Recently, many works were published on the topic of exploration planning for UAVs, but most of them assume that the UAV possesses a reliable state estimator, a depth sensor, and sufficient computational capabilities \cite{sonugur2023review}. However, in our work, we aim to extend the current state-of-the-art approach for autonomous multi-UAV exploration to consumer-level UAVs, such as the DJI Mini 3 Pro.
These only provide us with monocular images, and velocity and orientation measurements of limited precision ($\pm\SI{0.1}{\meter\per\second}$, $\pm\SI{0.1}{\degree}$).

Our pipeline consists of state and depth estimators, as well as an exploration planner, which extends the baseline to meet the limitations of these odometry and depth estimators.
Our main modifications are connected with the viewpoint pairs selection process and with the path planning through them. After we ensure that our single-UAV exploration with consumer-grade DJI drones is successful, we extend our approach to a multi-UAV case. The pipeline for the multi-UAV exploration consists of several UAVs that perform the exploration independently from each other, with the planner from the single-UAV approach. However, we add the base node, which is responsible for coordinating UAVs, accumulating data about the explored space, and assigning workload.

The contributions of the paper are as follows:
\begin{itemize}
    \item We propose a method of sampling and evaluating viewpoint pairs to produce images suitable for depth estimation. Additionally, we adapt the hierarchical exploration planner to viewpoint pairs instead of single viewpoints.
    \item We modify the trajectory planner to satisfy motion constraints by introducing a yaw-aware A\textsuperscript{*} planner and a joint optimization step with rotation constraints. 
    \item We introduce a semi-distributed multi-UAV exploration method, where the base node handles global planning and map updates due to the limited computational capabilities of consumer-grade UAVs.
    
\end{itemize}

\section{Related Works}

Exploration of unknown environments is a fundamental task, and one of the first steps is determining how to choose the best viewpoints. Most classical approaches can be divided into three groups: sampling-based, frontier-based, and a combination of the two. Sampling-based approaches randomly generate viewpoints and then evaluate them based on the information gain they provide, as in \cite{papachristos2017uncertainty}, \cite{dang2020graph}, and \cite{schmid2020efficient}. This approach offers a way to directly determine the next best view, but it is computationally demanding. Frontier-based approach, on the other hand, is less computationally demanding, but does not take into account the information gain, and was used in \cite{julia2012comparison}, \cite{cieslewski2017rapid}, and \cite{deng2020frontier}. Many methods, such as  \cite{heng2014autonomous}, \cite{heng2015efficient}, and \cite{dai2020fast}, combine frontier-based and sampling-based approaches to trade off computation time and information gain.

FUEL \cite{zhou2021fuel} is the pipeline for autonomous UAV exploration of the unknown environment. Their primary focus was on maintaining the high exploration rate while ensuring safe and energy-efficient movement of the UAV. The pipeline consists of two main parts: an incremental frontier information structure and a hierarchical planner, as well as a mapping module and a motion controller.

The authors introduce a concept of frontier information structure (FIS), which contains all the necessary information for exploration and can quickly be updated incrementally when new information arrives. After a new sensor measurement is received, they update the information about affected frontiers. The viewpoint poses are sampled in the cylindrical coordinate system around the average, and evaluated based on the coverage. The connection cost between each pair of clusters denotes the time lower bound when moving between the best viewpoints of each frontier. The motion is planned in a coarse-to-fine manner. Firstly, they create a global path that passes through all unknown regions. Then they create and refine the local path that passes through the desired viewpoints to ensure high information gain and a smooth path. And finally, they generate a safe and dynamically optimal motion of the quadrotor for it to reach the first viewpoint.

Later, the authors of FUEL extended their approach to the multi-quadrotor case with RACER \cite{zhou2023racer}. In their work, they improved their autonomous exploration pipeline and introduced a communication protocol for multiple UAVs. As a result, they ensure that all quadrotors explore distinct regions with only limited asynchronous communication between them. 

The volumetric map is constantly subdivided into disjoint hgrid cells that are used as elementary task units for UAVs to explore. The main difference between the previous exploration planner and the new one is the addition of a global coverage path, which passes through all unexplored hgrid cells assigned to the robot. Cells are distributed between UAVs via pairwise interactions to minimize the length of coverage paths through them. Additionally, UAVs exchange not only their allocated grids, but also the trajectories and explored map chunks.

\section{Exploration Using Consumer-grade UAVs}

\subsection{Single-UAV Exploration Method}

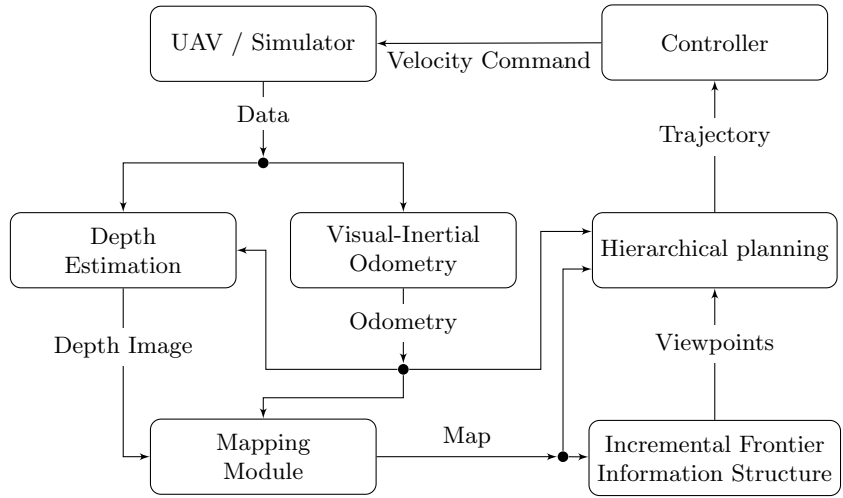
\begin{figure}[h]
    \vspace{-1em}
    \centering
    \begin{tikzpicture}[auto, 
    node distance = 10mm and 6mm,
    N/.style = {draw, fill=white, rectangle, rounded corners, minimum height=10mm, minimum width=30mm, align=center},
        pin/.style= {circle, fill,inner sep=0.5mm, align=center},
    >=latex'
    ]
    \node[N] (uav)  {UAV / Simulator};
    \node[pin, below=of uav]  (data) {};
    \node[N, below left=6mm and 3mm of data]  (depth)    {Depth \\ Estimation};
    \node[N, below right=6mm and 3mm of data]  (vio) {Visual-Inertial \\ Odometry};
    \node[pin, below=of vio]  (odometry) {};
    \node[N, below left=6mm and 3mm of odometry]  (mapping) {Mapping \\Module};
    \node[pin, right=24 mm of mapping]  (map) {};
    \node[N, right=10 mm of vio]  (plan) {Hierarchical planning};
    \node[N, below=17.5 mm of plan]  (fis) {Incremental Frontier \\ Information Structure};
    \node[N, above=17.5 mm of plan ]  (controller) {Controller};
    \draw [->] (controller) -- node{Velocity Command} (uav);
    \draw [->] (uav.south) -| node[below=2mm, fill=white]{Data} (data.north);
    \draw [->] (data) -| (depth);
    \draw [->] (data) -| (vio);
    \draw [->] (vio.south) -| node[below=2mm, fill=white]{Odometry} (odometry);
    \draw [->] (odometry.west) -- ([xshift= -17.5mm]odometry.west) |- (depth.east);
    \draw [->] (odometry.south) |- ([yshift= -3mm, xshift= -5mm]odometry.south) -| (mapping);
    \draw [->] (odometry.east) -- ([xshift=17.5mm]odometry.east) |- ([yshift= 2.5mm]plan.west);
    \draw [->] (depth) |- node[above=12mm, fill=white]{Depth Image} (mapping);
    \draw [->] (mapping) -- node{Map} (map);
    \draw [->] (map) -- (fis);
    \draw [->] (map) |- ([yshift= -2.5mm]plan.west);
    \draw [->] (fis.north) -| node[above=7.5mm, fill=white]{Viewpoints} (plan.south);
    \draw [->] (plan.north) -| node[above=7.5mm, fill=white]{Trajectory} (controller.south);
    \end{tikzpicture}
    \caption{Overview of the proposed pipeline for single-UAV exploration.}
    \label{fig:pipeline_single}
    \vspace{-2em}

\end{figure}

Our autonomous exploration method is presented in \Cref{fig:pipeline_single}. 
The first part of the pipeline consists of state and depth estimators that process the UAV's sensor data to produce the current odometry and depth image required for the exploration planner. In the second part, the exploration planner creates and updates the map, identifies frontiers and picks viewpoints, plans the trajectory, and sends velocity commands. 
All of these steps are performed on a ground control station, since the consumer-grade UAVs do not have sufficient onboard computing capabilities.
In case of DJI UAVs, the generated velocity commands can be executed via the Mobile SDK as used in \cite{vio}.

As the state estimator for our model, we use the visual-inertial odometry estimation model, which was proposed in \cite{vio}. The mapping module computes both the volumetric occupancy map and the Signed Euclidean Distance Field (ESDF) as proposed in \cite{han2019fiesta}, as well as performs hgrid decomposition. The controller includes the feedback loop to account for the difference between sent and executed velocity commands, and its output is rounded to the resolution of the velocity readings. 

\subsubsection{Depth Estimation}
To avoid revisiting frontiers and to create a high-quality map, we decided not to use fast depth estimation pipelines that produce only sparse depth maps. Instead, we use the transformer-based deep learning depth estimator MASt3R \cite{leroy2024grounding}, which produces dense depth maps. 
The model only accepts images as input and approximates camera intrinsics and extrinsics, which can result in scaling errors. 
However, since camera poses and parameters are known to us, we extract dense image correspondences before they are passed to the model aligner. 
From these, we estimate the corresponding 3D positions using the VIO poses and iterative DLT triangulation.
Since depth estimation requires multiple different viewpoints, we can either send images to the depth estimator uniformly after some time has passed, or explicitly plan at least two views per frontier.

\subsubsection{Viewpoint Planning} 
The Frontier Information Structure (FIS) stores and incrementally updates information about frontiers, using principal component analysis to estimate their orientations.
In the sampling step, we search for a pair of viewpoints from which the frontier can be reconstructed. 
We sample candidate poses uniformly around the edges of the frontier, with the yaw sampled around the direction towards the center. 
After filtering for sufficient FoV coverage, pairs of viewpoints $V_1, V_2$ are evaluated based on a heuristic reconstructability score $h$, which is inspired by \cite{smith2018aerial}.
It does not only consider overlap between the observed voxels $C_1, C_2$ but also maximal distance $d$, parallax $\alpha$, and observation angles $\theta_1,\theta_2$:
\begin{equation}
    h(V_1, V_2, C_1, C_2) = w_1(\alpha) w_2(d) w_3(\alpha) w_4(\theta) \frac{|C_1\cap C_2|}{|C_1\cup C_2|}.
\end{equation}
The parallax dependency of triangulation error and multi-view matchability are modeled as
\begin{equation}
    w_1(\alpha)=\frac{1}{1+e^{-k_1(\alpha - \alpha_\text{min})}}\quad \text{ and }\quad w_3(\alpha)=1-\frac{1}{1+e^{-k_3(\alpha - \alpha_\text{max})}},
\end{equation}
with $k_1, k_3$ determining the slope, $\alpha_\text{min}$ the minimal parallax for successful triangulation and $\alpha_\text{max}$ the maximal parallax for successful stereo matching.
Additionally, $w_2$ favors close-up views and $w_4$ disfavors shallow observation angles:
\begin{equation}
    w_2(d) = 1 - \min(1,\frac{d}{d_\text{max}}), \quad w_4(\theta_1,\theta_2) = \min(||\cos(\theta_1)||,||\cos(\theta_2)||),
\end{equation}
where $d_\text{max}$ is the maximal desired distance that depends on ray length limitations in the mapping module.
After numerous tests, we chose parameters to be $k_1 = 32$, $k_3 = 8$, $\alpha_\text{min} = \frac{\pi}{8}$, $\alpha_\text{max} = \frac{\pi}{3}$ and $d_\text{max} = \SI{3.5}{\meter}$.

Close view poses with significantly different yaws are difficult to connect with rotation-constrained trajectories.
Thus, we first sample viewpoints on lines parallel to the frontier.
The corresponding yaws are chosen such that the frontier and line are both covered by the field of view (FoV).
Thus, they can be traversed without rapid rotations.
Only if no pair with a sufficient reconstruction heuristic is found, do we fall back to uniform sampling. 

\subsubsection{Trajectory Planning} 
Exploration planning is divided hierarchically into three steps in a coarse-to-fine manner: global path planning, local viewpoint refinement, and trajectory generation.
In global planning, a coverage path through all frontiers is generated by solving the Traveling Salesman Problem.
To keep the search space small, we do not consider each viewpoint individually but represent each viewpoint pair by its midpoint.
As in FUEL, connection costs are estimated as time lower bounds when moving between two frontiers.
When restricting flight directions to the FoV, this significantly underestimates rotation costs.
Thus, we additionally insert an intermediate yaw point $\gamma_k = \arctan\frac{(p_j-p_i)_y}{(p_j-p_i)_x}$.

After the global path is found, we locally refine viewpoints.
Here, we have to refine viewpoints in one pair jointly because changing a viewpoint in one pair independently of its partner may result in a significant decrease in reconstructability.
Thus, we build a graph of viewpoint candidate nodes, where each viewpoint is only connected to its partner and to viewpoints corresponding to other frontiers. 
We then find the shortest path in the graph, starting from the current position, passing through the viewpoints, and ending in the midpoint of the last frontier in the local path window.

In contrast to the baseline trajectory generator, where position and yaw trajectories are created separately, we added a joint optimization step and considered more rotational constraints to avoid losing sight of visual features. Keeping the yaw aligned with the heading of the robot is vital for safe exploration. We introduce a limited FoV-constraint inspired by \cite{liu2018towards}, which penalizes the difference between yaw and the angle of the velocity vector. The result of the optimization procedure heavily depends on the initialization, so we propose to use a yaw-aware A\textsuperscript{*} planner. In addition to adding yaw to the state space, we consider two constraints in the node expansion step. Firstly, we want to avoid rotations on the spot. To do so, we check that there is a non-zero translational step between the current point and the neighbor. Secondly, we make sure that the velocity vector between the current pose and the neighbor is always within the FoV.

\subsection{Multi-UAV Exploration Method}
In the baseline, the authors established a pairwise request-response communication protocol, enabling UAVs to update the assigned workload with only limited communication. The assigned cells are redistributed between the pair of robots to minimize the length of the coverage path, subject to capacity constraints. All robots also continuously advertise their trajectories and explored map chunks. Since consumer-grade UAVs do not possess the computational power and receive commands from the controller or a ground station, we propose to use a semi-distributed communication scheme. Our pipeline for semi-distributed multi-UAV exploration consists of a single base node and multiple UAVs that perform exploration of individual areas, similar to a single UAV pipeline. 

The base node is responsible for three main things. It creates the global map from the positions of UAVs and depth images, and sends the map and the frontier list to UAVs. Secondly, when the UAV has visited the viewpoint, it sends its position and image to the base node, which accumulates them and communicates with the depth estimator. Finally, to avoid suboptimal pairwise hgrid reassignment, it performs the workload assignment and global planning for all UAVs in one iteration.

\section{Experiments}
We assess the performance of our model compared to the baseline when used with consumer-grade UAVs.
The evaluation is done in a simulation.
We use Flightmare \cite{song2020flightmare}, which is a modular simulator that features a rendering engine built on Unity, capable of outputting RGB and depth images and odometry for multiple quadrotors, and visualizing them in the chosen environment. We provide results of exploring the outdoor industrial environment with a size of 15 x 15 x 3 m. We tested out our pipeline in different scenes, but chose this one because it contains the following challenging parts: a homogeneous open area, a narrow gate, obstacles of different heights, such as crates, and an inaccessible region inside a big pipe. 
To evaluate the quality of depth estimation, we collect reconstruction accuracy (ACC), surface coverage (COV), and average Hausdorff distance (AHD) metrics. To evaluate the quality of depth estimation, we collect relative pose error (RPE) and absolute trajectory error (ATE).

    \begin{table}[!htb]
        \vspace{-1em}
        \centering
        \begin{tabular}{c|ccc}
            \toprule
            \textbf{method} & \textbf{AHD [m] \textdownarrow} & \textbf{ACC \textuparrow} & \textbf{COV \textuparrow} \\ \midrule
             no & $0.68 \pm 0.09$ & $0.53 \pm 0.02$ & $0.39 \pm 0.06$ \\ 
             overlap & $0.41 \pm 0.05$ & $0.74 \pm 0.06$ & $0.59 \pm 0.05$ \\
             heuristic & $\mathbf{0.30 \pm 0.02}$ & $\mathbf{0.93 \pm 0.02}$ & $\mathbf{0.69 \pm 0.02}$ \\
            \bottomrule
        \end{tabular}
        \caption{Map quality metrics for viewpoint pairs evaluation methods.}
        \label{tab:FIS_sort}
        \vspace{-3em}
    \end{table}

\begin{figure}[!htb]
        \vspace{-1em}
        \centering
        \subfloat[\centering no]{{\includegraphics[width=.3\linewidth]{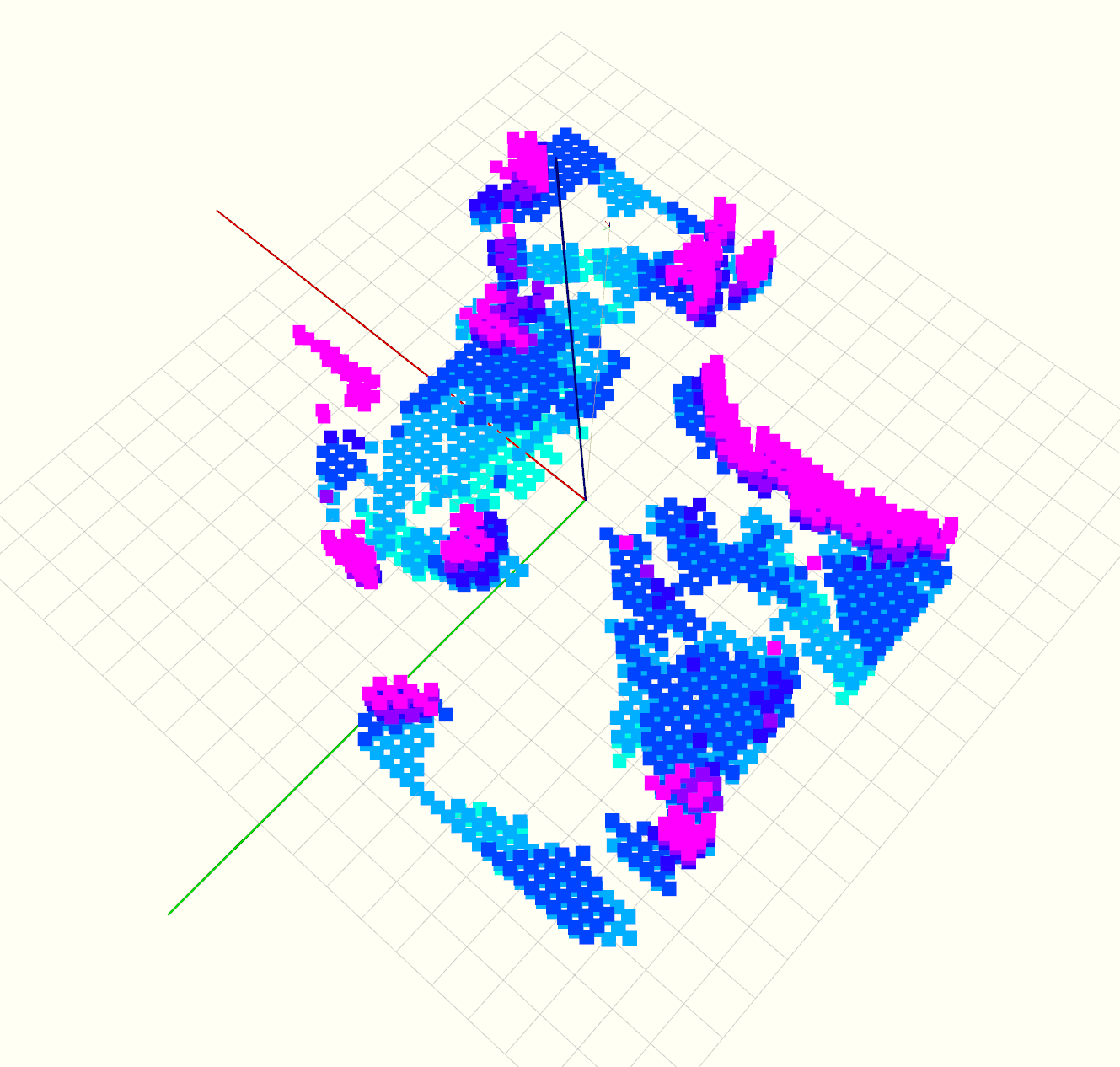}}}
        \subfloat[\centering overlap]{{\includegraphics[width=.3\linewidth]{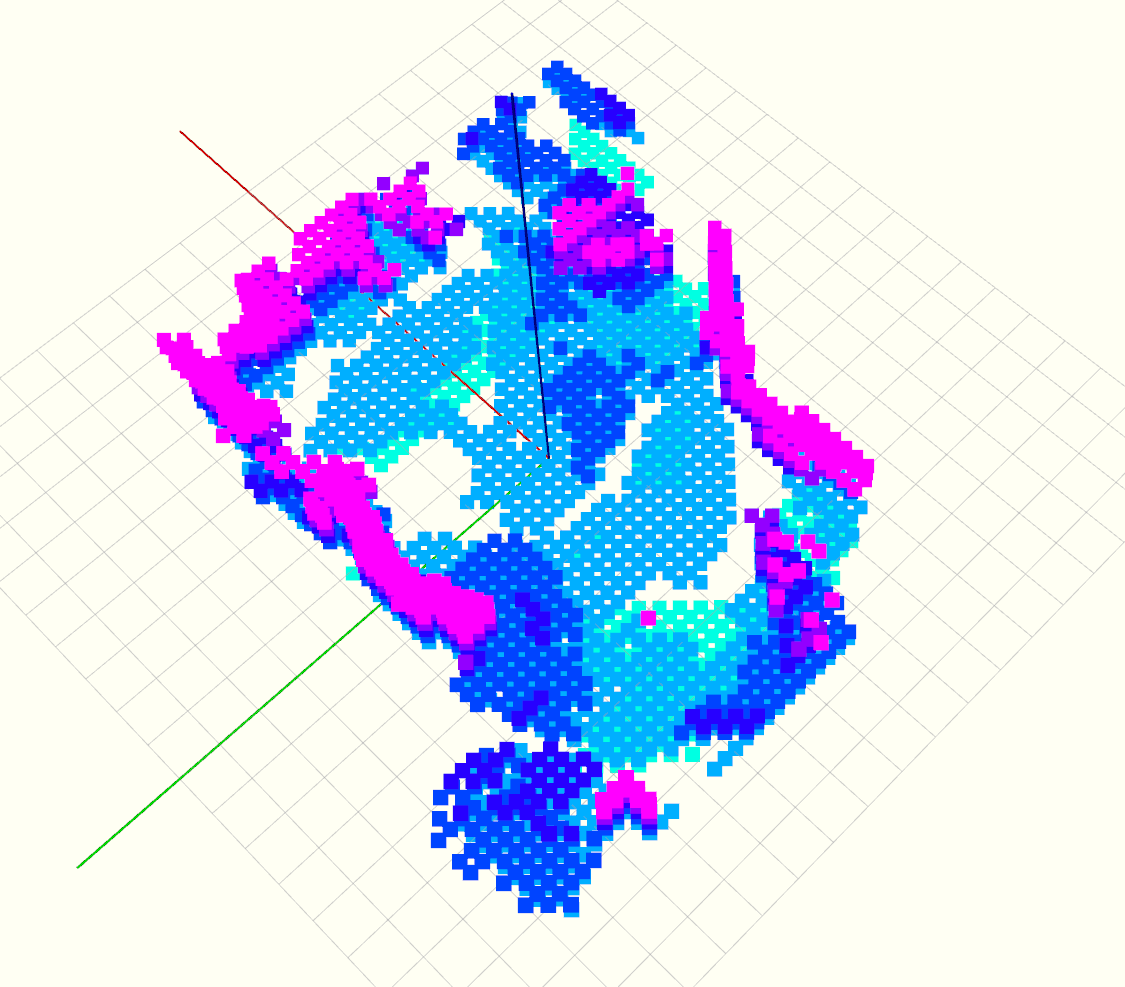}}}
        \subfloat[\centering heuristic]{{\includegraphics[width=.3\linewidth]{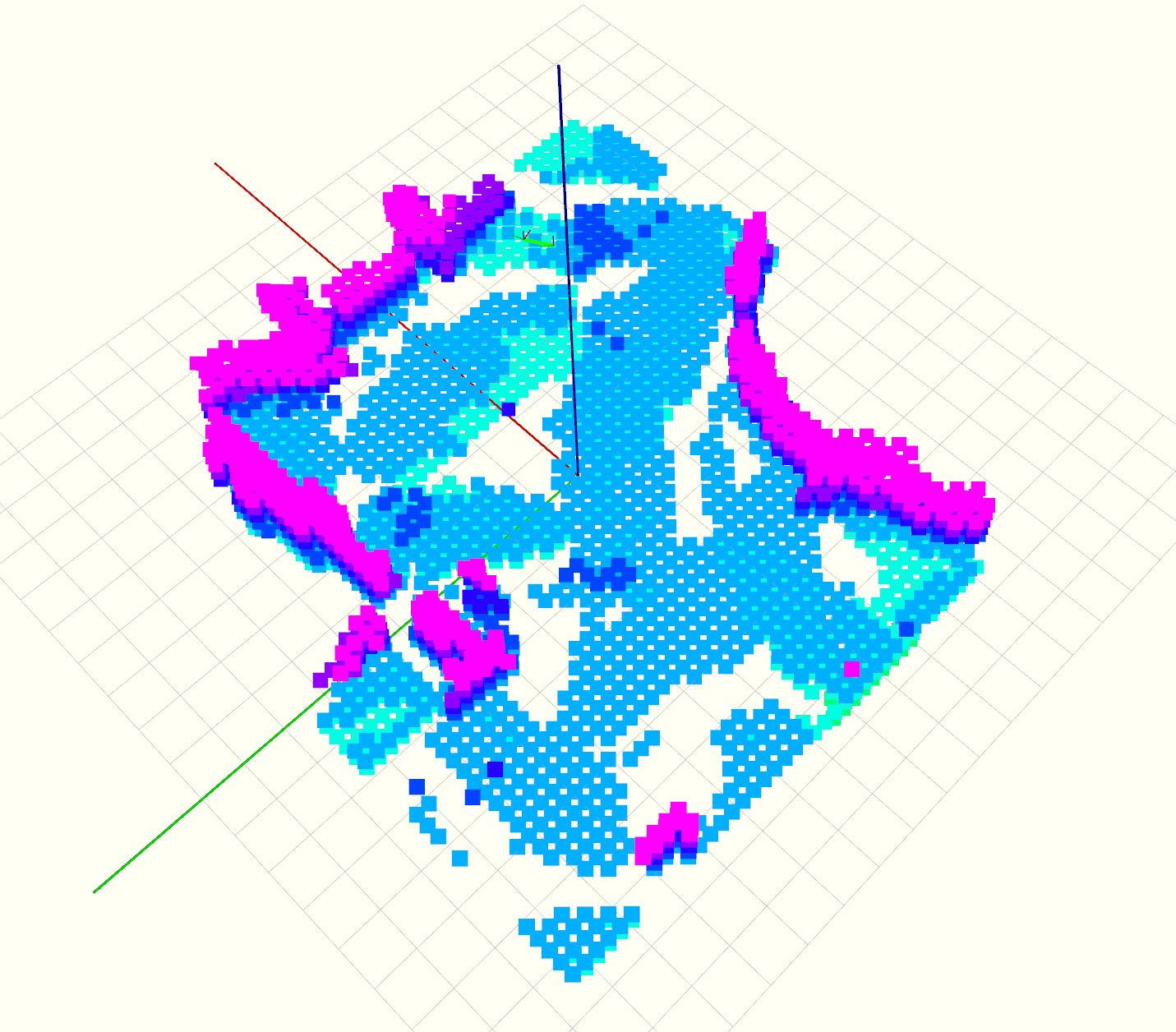}}}
        \caption{Map for different viewpoint pairs evaluation methods.}
        \label{fig:FIS_sort_map}
        \vspace{-2em}
    \end{figure}

First, we test the effect of our modifications that we proposed in the Viewpoint Planning section for viewpoint pairs sampling and evaluation. We compare the reconstructed map from images sent uniformly every 0.5 m with the one from images sent at viewpoint pairs, where viewpoints were sampled uniformly around edges of the frontier. Additionally, we test the influence of the heuristic-based pair evaluation method on the reconstructed map compared to the basic overlap-based method. Resulting metrics are presented in \Cref{tab:FIS_sort} and the map is shown in \Cref{fig:FIS_sort_map}. Viewpoint pair sampling increased the quality of depth estimation because it ensures that the overlap between images is sufficient and viewpoints are located far enough away from each other for the triangulation to work. Additionally, the heuristic-based method significantly outperforms the overlap-based method, because it considers the difference in the observation angles and avoids motion perpendicular to the frontier. 
    
Finally, we show the influence of line sampling on the map and VIO quality in \Cref{tab:FIS_sample_map}, and on the path in \Cref{tab:FIS_sample_path}. As we can see, the inclusion of line sampling leads to longer paths with fewer rotations, which increases the quality of VIO. In all experiments below, we used the combined sampling approach, which trades off the path length and viewpoint pairs quality with fewer rotations and VIO quality.  

    \begin{table}[!htb]
        \vspace{-1em}
        \centering
        \begin{tabular}{c|ccc|cc}
            \toprule
            \textbf{method} & \textbf{AHD [m] \textdownarrow} & \textbf{ACC \textuparrow} & \textbf{COV \textuparrow} & \textbf{RPE [m] \textdownarrow} & \textbf{ATE [m] \textdownarrow}\\ \midrule 
             uniform & $\mathbf{0.32 \pm 0.02}$ & $\mathbf{0.92 \pm 0.02}$ & $\mathbf{0.67 \pm 0.03}$ & $0.02 \pm 0.01$ & $0.17 \pm 0.04$\\
             line & $0.39 \pm 0.05$ & $0.87 \pm 0.05$ & $0.58 \pm 0.02$ & $\mathbf{0.01 \pm 0.00}$ & $\mathbf{0.11 \pm 0.06}$\\
             combined & $0.33 \pm 0.02$ & $0.87 \pm 0.02$ & $0.66 \pm 0.01$ & $\mathbf{0.01 \pm 0.00}$ & $0.12 \pm 0.04$ \\
            \bottomrule
        \end{tabular}
        \caption{Map and VIO quality metrics for viewpoint pairs sampling methods.}
        \label{tab:FIS_sample_map}
        \vspace{-4em}
    \end{table}
    
    \begin{table}[!htb]
        \vspace{-1em}
        \centering
        \begin{tabular}{c|cccc}
            \toprule
            \textbf{method} & \textbf{time [s] \textdownarrow} & \textbf{length [m] \textdownarrow} & \textbf{avg. velocity [$\frac{m}{s}$] \textuparrow} & \textbf{energy [$\frac{m^2}{s^2}$] \textdownarrow}\\ \midrule
            uniform & $\mathbf{567 \pm 5}$ & $\mathbf{161.9 \pm 5.2}$ & $0.29 \pm 0.01$ & $\mathbf{50.37 \pm 4.15}$  \\
            line & $639 \pm 48$ & $205.6 \pm 16.5$ & $\mathbf{0.32 \pm 0.01}$ & $79.02 \pm 5.21$ \\
            combined & $663 \pm 19$ & $199.1 \pm 6.2$ & $0.30 \pm 0.01$ & $67.69 \pm 1.47$ \\
            \bottomrule
        \end{tabular}
        \caption{Path metrics for viewpoint pairs sampling methods.}
        \label{tab:FIS_sample_path}
        \vspace{-3em}
    \end{table}
    
\begin{figure}[!htb]
        \vspace{-1em}
        \centering
        \subfloat[\centering baseline]{{\includegraphics[width=.3\linewidth]{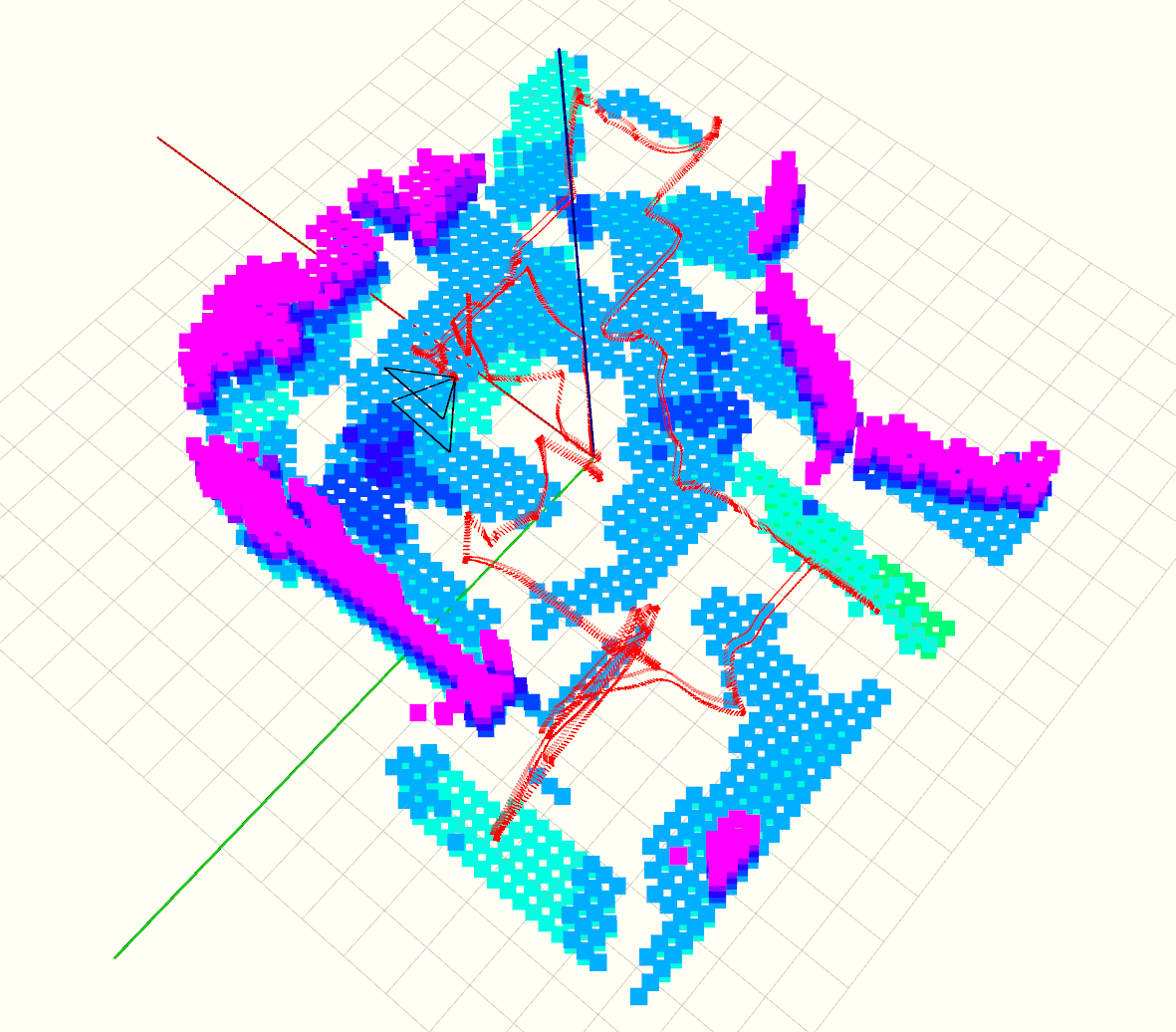}}}
        \subfloat[\centering constrained]{{\includegraphics[width=.3\linewidth]{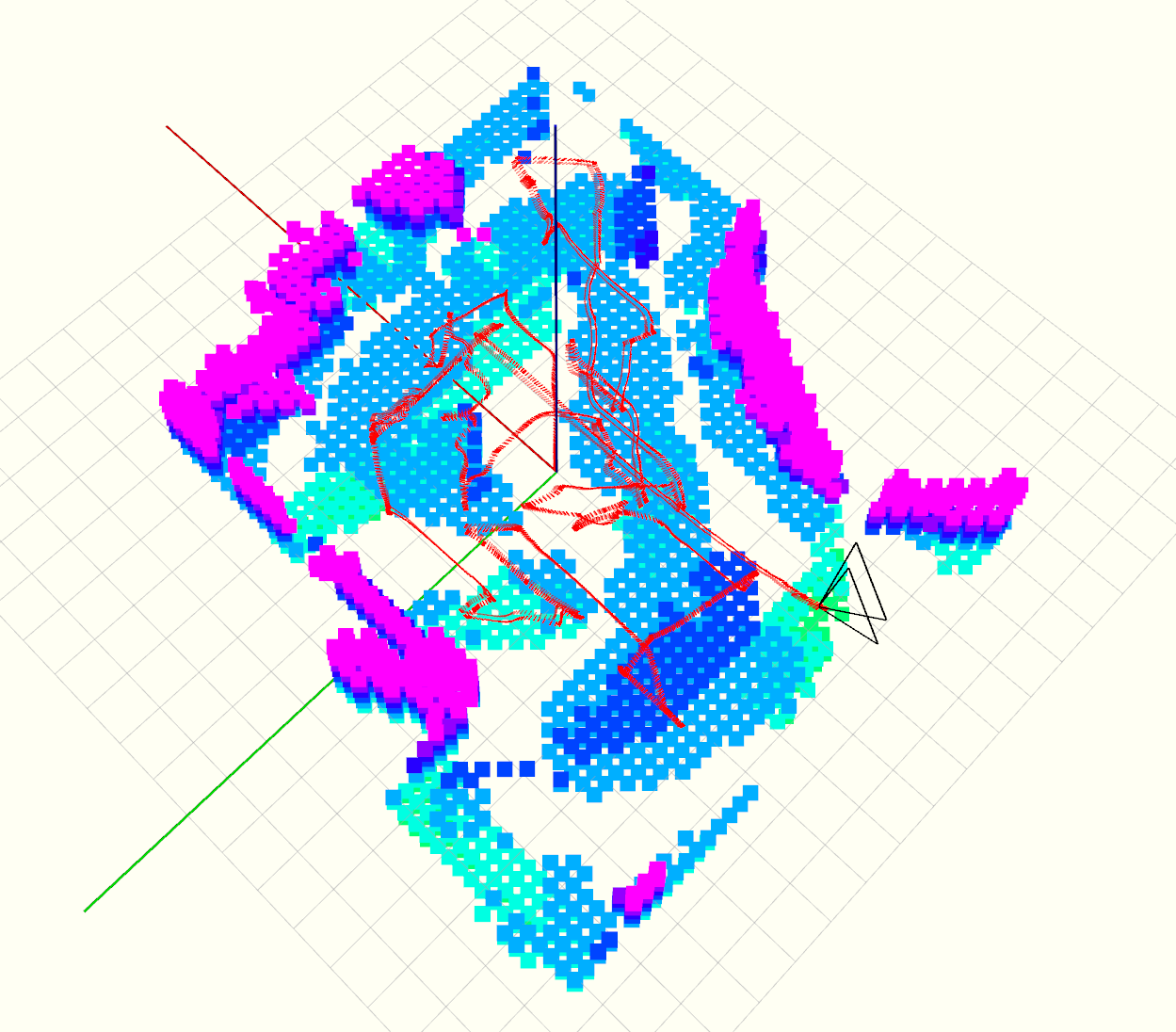}}}
        \subfloat[\centering yaw-aware A\textsuperscript{*}]{{\includegraphics[width=.3\linewidth]{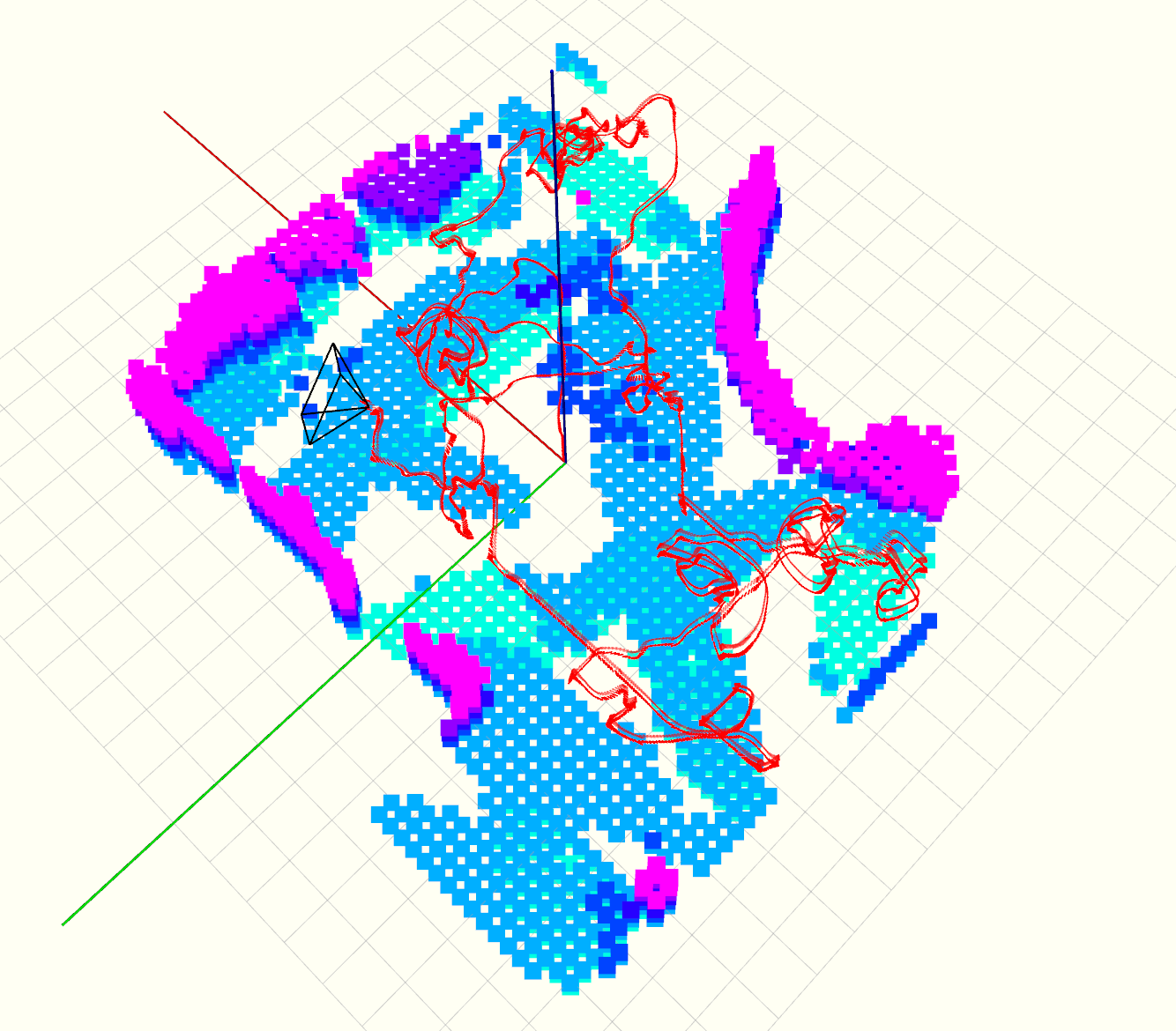}}}
        \caption{Map for trajectory generation methods.}
        \label{fig:traj_map}
        \vspace{-2em}
    \end{figure}

Next, we compare our yaw-aware A* trajectory generator with the updated and baseline soft-constrained approaches. You can see the estimated and ground truth trajectories in \Cref{fig:traj_map}, the path metrics in \Cref{tab:traj_path}, and map and odometry metrics in \Cref{tab:traj_map_odom}. Both baseline and updated soft-constrained methods fail to find the path between viewpoints within one pair that satisfies all desired constraints. A UAV travels on a straight line path with yaw perpendicular to the velocity vector, which leads to a shorter route and exploration time, but makes the trajectory unsafe. Our yaw-aware A\textsuperscript{*} solves this problem at the cost of longer path length and exploration time.

    \begin{table}[!htb]
        \vspace{-1em}
        \centering
        \begin{tabular}{c|cccc}
            \toprule
             \textbf{method} & \textbf{time [s] \textdownarrow} & \textbf{length [m] \textdownarrow} & \textbf{avg. velocity [$\frac{m}{s}$] \textuparrow} & \textbf{energy [$\frac{m^2}{s^2}$] \textdownarrow}\\ \midrule
            baseline & $420 \pm 110$ & $107.9 \pm 18.9$ & $0.27 \pm 0.03$ & $\mathbf{50.49 \pm 12.06}$  \\
            constrained & $\mathbf{372 \pm 37}$ & $\mathbf{101.6 \pm 6.3}$ & $0.28 \pm 0.03$ & $51.54 \pm 9.73$ \\
            A\textsuperscript{*} & $576 \pm 24$ & $171.7 \pm 10.2$ & $\mathbf{0.30 \pm 0.01}$ & $63.98 \pm 3.53$ \\
            \bottomrule
        \end{tabular}
        \caption{Path metrics for trajectory generation methods.}
        \label{tab:traj_path}
        \vspace{-0em}
    \end{table}
    
    \begin{table}[!htb]
        \vspace{-1em}
        \centering
        \begin{tabular}{c|ccc|cc}
            \toprule
            \textbf{method} & \textbf{AHD [m] \textdownarrow} & \textbf{ACC \textuparrow} & \textbf{COV \textuparrow} & \textbf{RPE [m] \textdownarrow} & \textbf{ATE [m] \textdownarrow}\\ \midrule
            baseline & $0.35 \pm 0.01$ & $0.86 \pm 0.02$ & $0.61 \pm 0.02$ & $\mathbf{0.01 \pm 0.00}$ & $0.18 \pm 0.06$\\
            constrained & $\mathbf{0.31 \pm 0.04}$ & $0.86 \pm 0.06$ & $\mathbf{0.66 \pm 0.05}$ & $\mathbf{0.01 \pm 0.00}$ & $0.26 \pm 0.09$\\
            A\textsuperscript{*} & $0.35 \pm 0.02$ & $\mathbf{0.92 \pm 0.01}$ & $0.64 \pm 0.02$ & $\mathbf{0.01 \pm 0.00}$ & $\mathbf{0.14 \pm 0.02}$\\
            \bottomrule
        \end{tabular}
        \caption{Map and VIO quality metrics for trajectory generation methods.}
        \label{tab:traj_map_odom}
        \vspace{-1em}
    \end{table}

\begin{figure}[!htb]
        \vspace{-1em}
        \centering
        \subfloat[\centering 2 UAVs]{{\includegraphics[width=.40\linewidth]{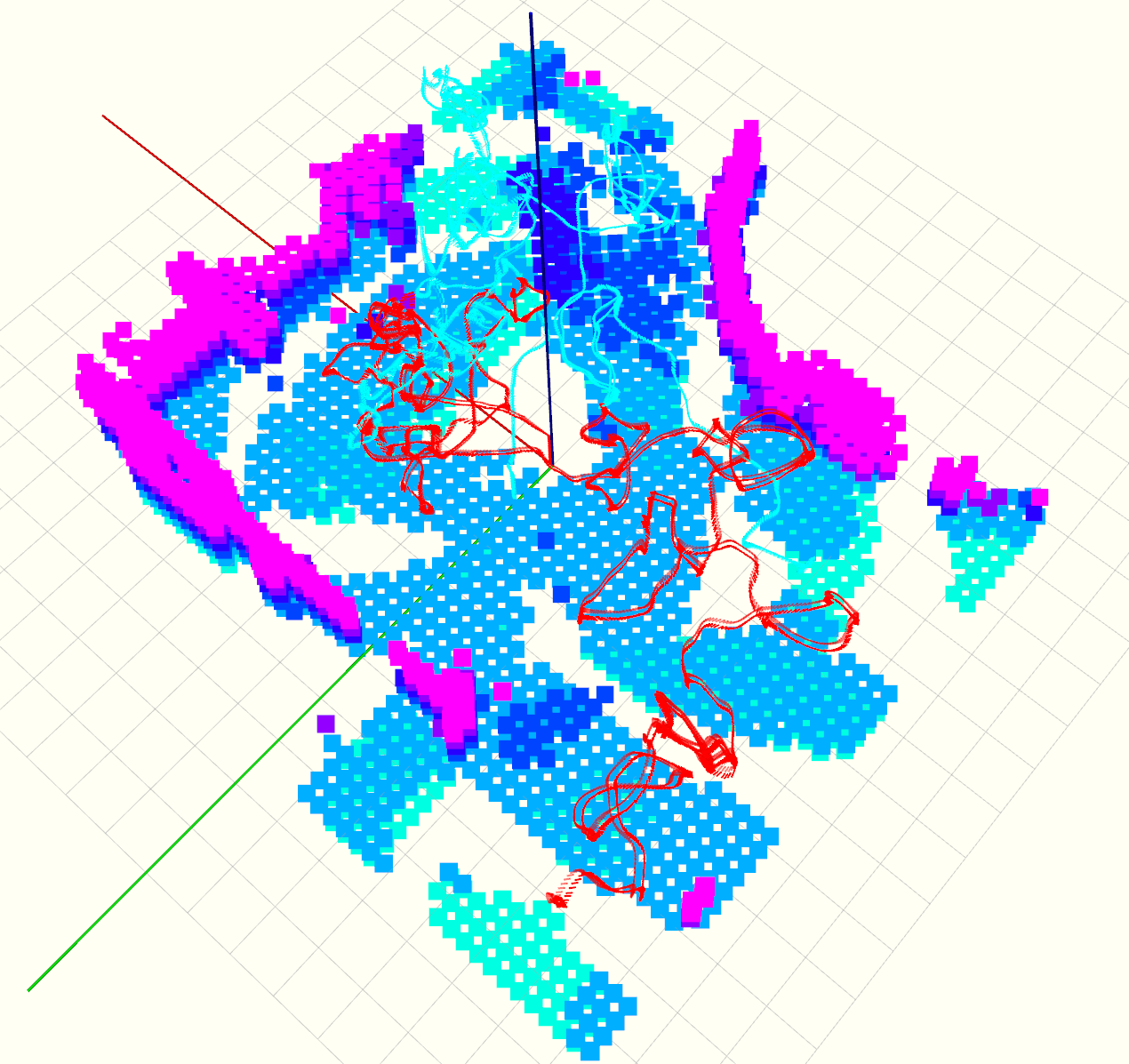}}}
        \subfloat[\centering 3 UAVs]{{\includegraphics[width=.40\linewidth]{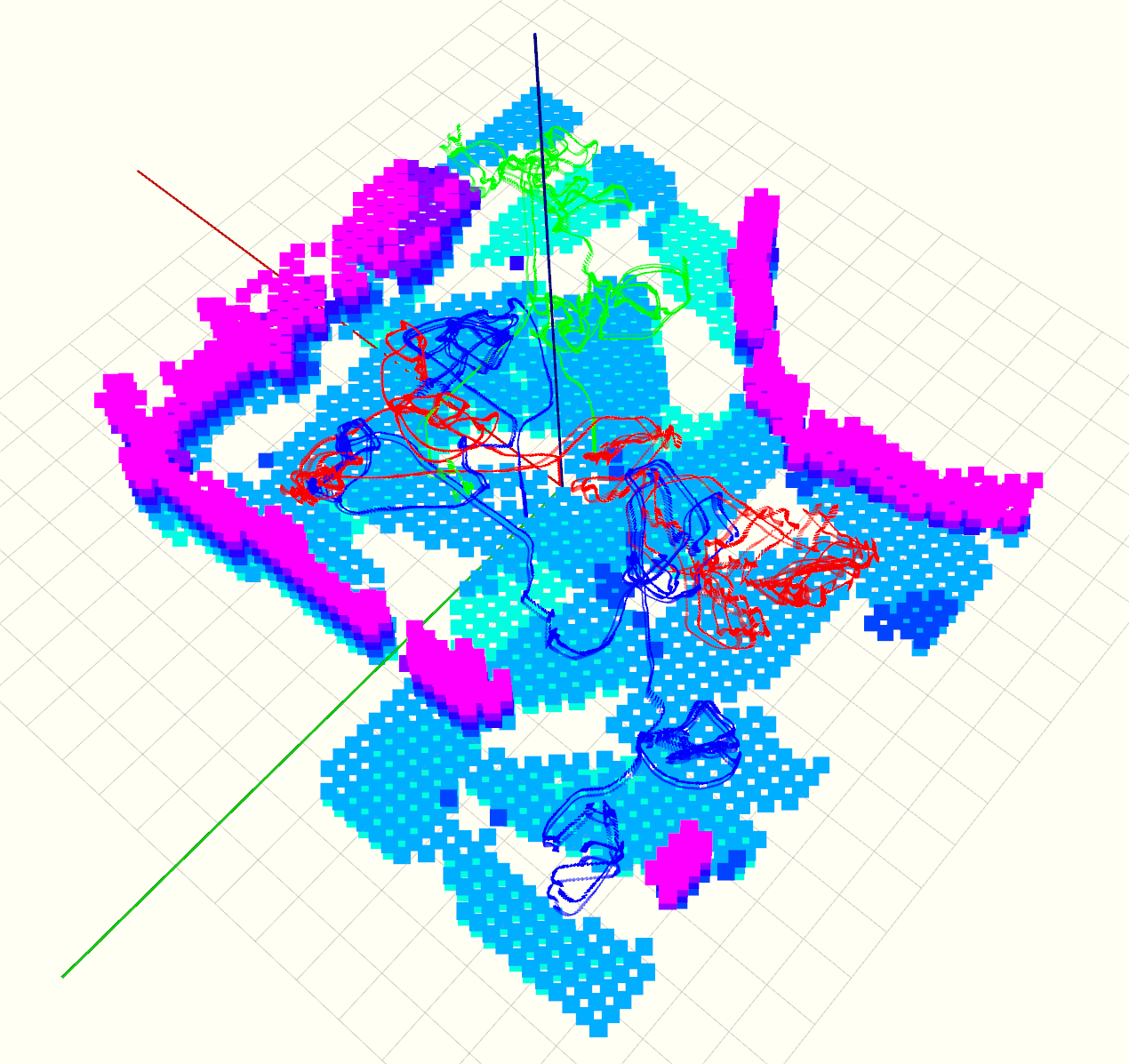}}}
        \caption{Map and path for semi-distributed exploration method and different numbers of UAVs.}
        \label{fig:multi_map}
        \vspace{-2em}
    \end{figure}

Finally, we evaluate the baseline multi-UAV exploration method with our exploration planner compared with our semi-distributed multi-UAV exploration method. Path metrics are presented in \Cref{tab:multi_path}, and reconstructed map and trajectories for different numbers of UAVs for our method are shown in \Cref{fig:multi_map}. Our method outperforms the baseline for both two and three UAVs due to a more optimal workload assignment, which is calculated for all UAVs in one iteration. The number of hgrid cells is limited, so calculating the coverage paths for all UAVs does not introduce a significant computational overhead.

    \begin{table}[!htb]
        \vspace{-1em}
        \centering
        \begin{tabular}{cc|cccc}
            \toprule
            \textbf{method} & \textbf{UAVs} & \textbf{time [s] \textdownarrow} & \textbf{length [m] \textdownarrow} & \textbf{avg. velocity [$\frac{m}{s}$] \textuparrow} & \textbf{energy [$\frac{m^2}{s^2}$] \textdownarrow}\\ \midrule
            single & 1 & $576 \pm 24$ & $171.7 \pm 10.2$ & $0.30 \pm 0.01$ & $63.98 \pm 3.53$ \\  \midrule
            baseline & 2 & $503 \pm 81$ & $147.0 \pm 24.2$ & $\mathbf{0.30 \pm 0.01}$ & $53.52 \pm 4.82$ \\
            our & 2 & $\mathbf{493 \pm 34}$ & $\mathbf{134.0 \pm 12.4}$ & $0.27 \pm 0.03$ & $\mathbf{47.40 \pm 4.34}$\\
            \midrule
            baseline & 3 & $426 \pm 47$ & $130.5 \pm 15.7$ & $\mathbf{0.31 \pm 0.01}$ & $49.06 \pm 5.79$\\
            our & 3 & $\mathbf{418 \pm 24}$ & $\mathbf{97.6 \pm 20.2}$ & $0.24 \pm 0.04$ & $\mathbf{37.24 \pm 8.14}$\\
            \bottomrule
        \end{tabular}
        \caption{Average path metrics for multi-UAV exploration methods. Length, avg. velocity, and energy are averaged per UAV. }
        \label{tab:multi_path}
        \vspace{-3em}
    \end{table}

\section{Conclusion}
In our work, we extended the baseline method for autonomous multi-UAV exploration to consumer-grade UAVs. To ensure the good quality of depth estimation, we proposed to use not viewpoints, but viewpoint pairs. We introduced the way to sample viewpoint pairs and evaluate their quality, and showed its influence on the reconstructed map. Odometry estimation and safety constraints require us to keep the direction of movement within the field-of-view of the UAV. To satisfy these constraints, we introduced the yaw-aware A\textsuperscript{*} planner and evaluated its performance compared to the baseline trajectory planner. Finally, due to the limited computational capabilities of consumer-grade UAVs, we proposed the semi-distributed communication scheme and showed that it outperforms the baseline fully-distributed method for different numbers of UAVs. 

Our method can be extended for larger exploration areas without any changes in the pipeline due to its distributed work assignment and the coarse-to-fine planning methods. However, in real-life experiments, the connection with the ground station may become unstable when exploring large regions. In future works, we would like to implement our pipeline for exploration of the real-world environment. Our main goal will be to test the scalability and generalizability of our method outside of the simulated scenario. 

\begin{footnotesize}
    \subsubsection{\ackname} This work has been supported by the German Federal Ministry of Research, Technology and Space (BMFTR) in the project “Kompetenzzentrum: Etablierung des Deutschen Rettungsrobotik-Zentrums (E-DRZ)”, grant 13N16477.
\end{footnotesize}

\bibliographystyle{styles/bibtex/spmpsci.bst}
\bibliography{bibfile_thesis}

@article{zhou2021fuel,
  title={Fuel: Fast uav exploration using incremental frontier structure and hierarchical planning},
  author={Zhou, Boyu and Zhang, Yichen and Chen, Xinyi and Shen, Shaojie},
  journal={IEEE Robotics and Automation Letters},
  volume={6},
  number={2},
  pages={779--786},
  year={2021},
  publisher={IEEE}
}

@article{zhou2023racer,
  title={Racer: Rapid collaborative exploration with a decentralized multi-uav system},
  author={Zhou, Boyu and Xu, Hao and Shen, Shaojie},
  journal={IEEE Transactions on Robotics},
  volume={39},
  number={3},
  pages={1816--1835},
  year={2023},
  publisher={IEEE}
}

@article{schmid2020efficient,
  title={An efficient sampling-based method for online informative path planning in unknown environments},
  author={Schmid, Lukas and Pantic, Michael and Khanna, Raghav and Ott, Lionel and Siegwart, Roland and Nieto, Juan},
  journal={IEEE Robotics and Automation Letters},
  volume={5},
  number={2},
  pages={1500--1507},
  year={2020},
  publisher={IEEE}
}

@article{dang2020graph,
  title={Graph-based subterranean exploration path planning using aerial and legged robots},
  author={Dang, Tung and Tranzatto, Marco and Khattak, Shehryar and Mascarich, Frank and Alexis, Kostas and Hutter, Marco},
  journal={Journal of Field Robotics},
  volume={37},
  number={8},
  pages={1363--1388},
  year={2020},
  publisher={Wiley Online Library}
}

@inproceedings{papachristos2017uncertainty,
  title={Uncertainty-aware receding horizon exploration and mapping using aerial robots},
  author={Papachristos, Christos and Khattak, Shehryar and Alexis, Kostas},
  booktitle={2017 IEEE international conference on robotics and automation (ICRA)},
  pages={4568--4575},
  year={2017},
  organization={IEEE}
}

@article{julia2012comparison,
  title={A comparison of path planning strategies for autonomous exploration and mapping of unknown environments},
  author={Juli{\'a}, Miguel and Gil, Arturo and Reinoso, Oscar},
  journal={Autonomous Robots},
  volume={33},
  pages={427--444},
  year={2012},
  publisher={Springer}
}

@inproceedings{cieslewski2017rapid,
  title={Rapid exploration with multi-rotors: A frontier selection method for high speed flight},
  author={Cieslewski, Titus and Kaufmann, Elia and Scaramuzza, Davide},
  booktitle={2017 IEEE/RSJ International Conference on Intelligent Robots and Systems (IROS)},
  pages={2135--2142},
  year={2017},
  organization={IEEE}
}

@article{deng2020frontier,
  title={Frontier-based automatic-differentiable information gain measure for robotic exploration of unknown 3D environments},
  author={Deng, Di and Xu, Zhefan and Zhao, Wenbo and Shimada, Kenji},
  journal={arXiv preprint arXiv:2011.05288},
  year={2020}
}

@article{heng2014autonomous,
  title={Autonomous visual mapping and exploration with a micro aerial vehicle},
  author={Heng, Lionel and Honegger, Dominik and Lee, Gim Hee and Meier, Lorenz and Tanskanen, Petri and Fraundorfer, Friedrich and Pollefeys, Marc},
  journal={Journal of Field Robotics},
  volume={31},
  number={4},
  pages={654--675},
  year={2014},
  publisher={Wiley Online Library}
}

@inproceedings{heng2015efficient,
  title={Efficient visual exploration and coverage with a micro aerial vehicle in unknown environments},
  author={Heng, Lionel and Gotovos, Alkis and Krause, Andreas and Pollefeys, Marc},
  booktitle={2015 IEEE International Conference on Robotics and Automation (ICRA)},
  pages={1071--1078},
  year={2015},
  organization={IEEE}
}

@inproceedings{dai2020fast,
  title={Fast frontier-based information-driven autonomous exploration with an mav},
  author={Dai, Anna and Papatheodorou, Sotiris and Funk, Nils and Tzoumanikas, Dimos and Leutenegger, Stefan},
  booktitle={2020 IEEE international conference on robotics and automation (ICRA)},
  pages={9570--9576},
  year={2020},
  organization={IEEE}
}

@article{liu2018towards,
  title={Towards search-based motion planning for micro aerial vehicles},
  author={Liu, Sikang and Mohta, Kartik and Atanasov, Nikolay and Kumar, Vijay},
  journal={arXiv preprint arXiv:1810.03071},
  year={2018}
}

@InProceedings{song2020flightmare,
  title= {Flightmare: A Flexible Quadrotor Simulator},
  author={Song, Yunlong and Naji, Selim and Kaufmann, Elia and Loquercio, Antonio and Scaramuzza, Davide},
  booktitle = {Proceedings of the 2020 Conference on Robot Learning},
  pages = {1147--1157},
  year = {2021}
}

@inproceedings{leroy2024grounding,
  title={Grounding image matching in 3d with mast3r},
  author={Leroy, Vincent and Cabon, Yohann and Revaud, J{\'e}r{\^o}me},
  booktitle={European Conference on Computer Vision},
  pages={71--91},
  year={2024},
  organization={Springer}
}

@article{smith2018aerial,
  title={Aerial path planning for urban scene reconstruction: A continuous optimization method and benchmark},
  author={Smith, Neil and Moehrle, Nils and Goesele, Michael and Heidrich, Wolfgang},
  year={2018},
  publisher={Association for Computing Machinery (ACM)}
}

@inproceedings{han2019fiesta,
  title={Fiesta: Fast incremental euclidean distance fields for online motion planning of aerial robots},
  author={Han, Luxin and Gao, Fei and Zhou, Boyu and Shen, Shaojie},
  booktitle={2019 IEEE/RSJ International Conference on Intelligent Robots and Systems (IROS)},
  pages={4423--4430},
  year={2019},
  organization={IEEE}
}

@article{vio,
  title={Autonomous Consumer-grade UAVs for Real-time Situational Awareness in GNSS-denied Search \& Rescue Operations},
  author={Schleich, Daniel and Quenzel, Jan and Behnke, Sven},
  journal={IEEE International Symposium on Safety, Security, and Rescue Robotics (SSRR)},
  year={2025}
}

@article{sonugur2023review,
  title={A Review of quadrotor UAV: Control and SLAM methodologies ranging from conventional to innovative approaches},
  author={Sonug{\"u}r, G{\"u}ray},
  journal={Robotics and Autonomous Systems},
  volume={161},
  pages={104342},
  year={2023},
  publisher={Elsevier}
}

\end{document}